\documentclass[letterpaper]{article} 
\usepackage{aaai25}  
\usepackage{times}  
\usepackage{helvet}  
\usepackage{courier}  
\usepackage[hyphens]{url}  
\usepackage{graphicx} 
\usepackage{natbib}  
\usepackage{caption} 
\usepackage{algorithm}
\usepackage{algorithmic}

\usepackage{amsfonts,amssymb}
\usepackage{soul}
\usepackage{color,xcolor}
\usepackage{tabularx}
\usepackage{amsmath}
\usepackage{bbding}
\usepackage{booktabs}
\usepackage{pifont} 
\usepackage{fontawesome}
\newcommand{\cmark}{\ding{51}}
\newcommand{\xmark}{\ding{55}}
\usepackage{makecell}
\usepackage{multirow}
\usepackage{multicol}

\usepackage{newfloat}
\usepackage{listings}
\DeclareCaptionStyle{ruled}{labelfont=normalfont,labelsep=colon,strut=off} 
\floatstyle{ruled}
\newfloat{listing}{tb}{lst}{}
\floatname{listing}{Listing}
\title{Bright-NeRF: Brightening Neural Radiance Field with Color Restoration from Low-light Raw Images}
\author{
    Min Wang\textsuperscript{\rm 1},
Xin Huang\textsuperscript{\rm 1},
Guoqing Zhou\textsuperscript{\rm 1},
Qifeng Guo\textsuperscript{\rm 2},
Qing Wang\textsuperscript{\rm 1}\thanks{Corresponding Author}
}
\affiliations{
    \textsuperscript{\rm 1}School of Computer Science, Northwestern Polytechnical University, Xi’an 710072, China \\
    \textsuperscript{\rm 2}Shenzhen Shenzhi Weilai Co., Ltd., Shenzhen 518000, China \\

    (wangmin\_phd, xinhuang)@mail.nwpu.edu.cn, 
   (zhouguoqing, qwang)@nwpu.edu.cn,
   guoqifeng001@gmail.com
}

\usepackage{bibentry}

\begin{document}

\maketitle

\begin{abstract}
Neural Radiance Fields (NeRFs) have demonstrated prominent performance in novel view synthesis. However, their input heavily relies on image acquisition under normal light conditions, making it challenging to learn accurate scene representation in low-light environments where images typically exhibit significant noise and severe color distortion. To address these challenges, we propose a novel approach, \textbf{Bright-NeRF}, which learns enhanced and high-quality radiance fields from multi-view low-light raw images in an unsupervised manner. Our method simultaneously achieves color restoration, denoising, and enhanced novel view synthesis. Specifically, we leverage a physically-inspired model of the sensor's response to illumination and introduce a chromatic adaptation loss to constrain the learning of response, enabling consistent color perception of objects regardless of lighting conditions. We further utilize the raw data's properties to expose the scene's intensity automatically. Additionally, we have collected a multi-view low-light raw image dataset to advance research in this field. Experimental results demonstrate that our proposed method significantly outperforms existing 2D and 3D approaches. Our code and dataset will be made publicly available.
\end{abstract}

%

\section{Introduction}

Neural Radiance Fields (NeRFs) have revolutionized the field of novel view synthesis by learning implicit scene representations from multi-view images, enabling the generation of high-quality 
views. Recently, 
NeRF has shown great potential in various applications such as relighting \citep{zhang2021nerfactor, rudnev2022nerf}, editing \citep{yuan2022nerf, song2023blending}, autonomous driving \citep{tonderski2024neurad}, robot navigation \citep{adamkiewicz2022vision} and object detection \citep{xu2023nerf, hu2023nerf}, 
demonstrating its versatility and robustness. However, the NeRF framework predominantly relies on multi-view images captured under well-lit conditions. In low-light environments, NeRF struggles to reconstruct scenes accurately due to inherent visual quality deficiencies such as low visibility, poor contrast, and significant noise, which severely impair its performance. Addressing these challenges is crucial for extending NeRF’s applicability to diverse real-world scenarios.

\begin{figure}[t]
\centering
\includegraphics[width=1.0\columnwidth]{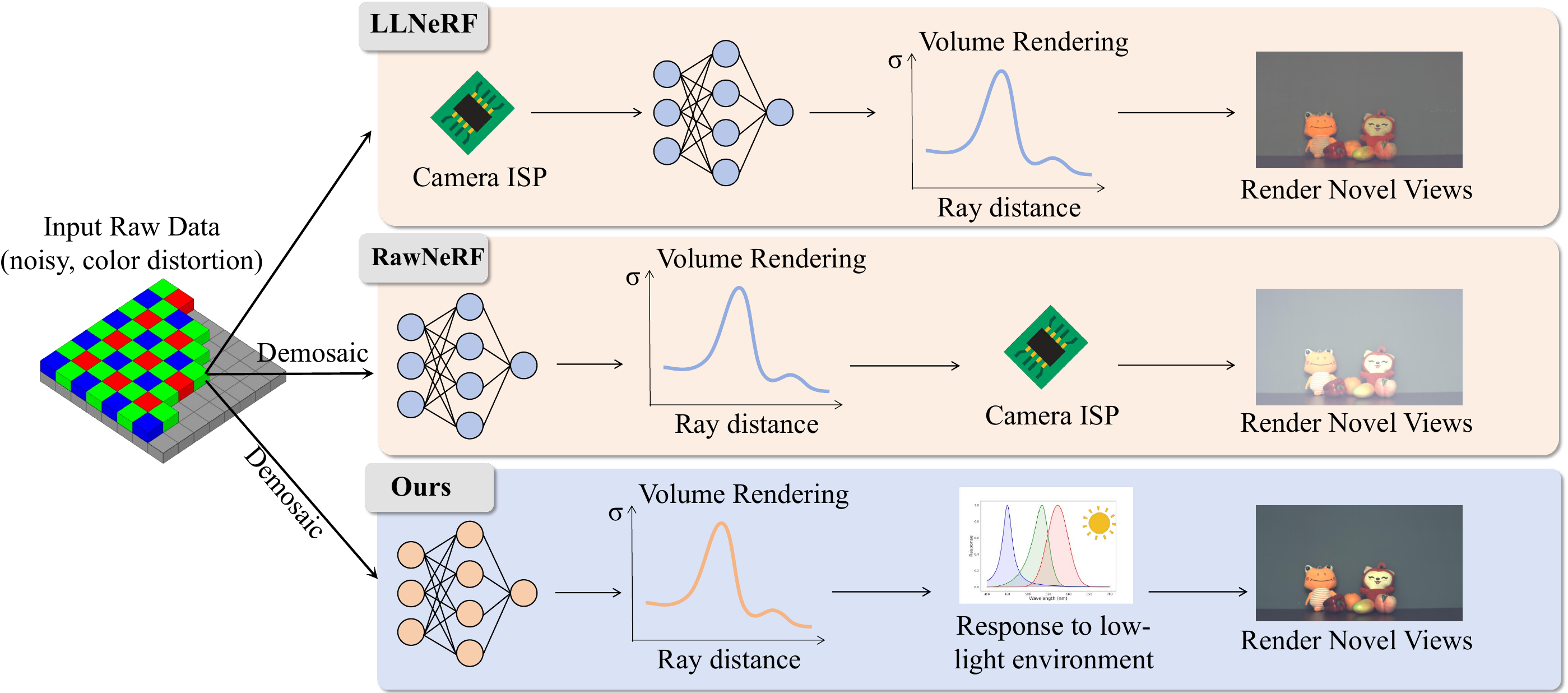} 
\caption{{\textbf{Comparison of LLNeRF, RawNeRF, and our Bright-NeRF.}} LLNeRF is trained on ISP-finished sRGB images, struggling to recover accurate color in extremely dark conditions via the low bit-depth sRGB data. Utilizing high bit-depth linear raw data, RawNeRF relies on a pre-calibrated ISP for post-processing the rendered images, which involves a complex and costly calibration process. Using a generic ISP offers a compromise, but it fails to deliver high-quality results across different cameras and scenarios. In contrast, Bright-NeRF eliminates these limitations by estimating the adaptive sensor's response to low-photon lighting conditions, ensuring consistent color perception and producing vivid color.}
\label{Fig.1}
\end{figure}

Several NeRF variants have attempted to reconstruct 3D scenes from degraded inputs such as reconstructing HDR scene representation from LDR views \cite{huang2022hdr}, motion blur \citep{ma2022deblur, lee2023dp}, large illumination variations \cite{martin2021nerf}, and aliasing \citep{barron2021mip, barron2022mip}. Despite the prevalence of these conditions in real-world scenarios, novel view synthesis in low-light environments has not yet been adequately addressed by these methods. To achieve normal-light novel view synthesis from images captured in low-light conditions, an intuitive approach involves initially enhancing multi-view images using 2D low-light enhancement methods~\citep{guo2016lime, chen2018learning, ma2022toward, dong2022abandoning, jin2023dnf}, followed by learning a NeRF from the processed images. However, these 2D methods tend to overfit to the training data processed by a fixed ISP, leading to poor generalization across the images captured by different cameras. Most importantly, they typically process each image independently and do not ensure consistency across views during rendering (as shown in Fig. \ref{Fig.5}), resulting in bias-guided NeRF and inconsistent enhancement results.

Recently, some low-light image enhancement (LLIE) NeRF methods have been proposed. LLNeRF \cite{wang2023lighting} focuses on recovering scene representations under normal-light conditions from a set of low-light sRGB images by decomposing colors into lighting-related and reciprocal components. However, it has difficulty in accurately decomposing colors in extremely low-light conditions (as shown in Fig \ref{Fig.6}). Additionally, limited by the bit-depth of sRGB images, LLNeRF struggles to recover fine color details, leading to unnatural color and lighting in the rendered images. Compared to sRGB data, raw data's unique high bit-depth characteristic is advantageous for low-light enhancement tasks. Under extremely low-light conditions, the linear properties and high bit-depth of raw images facilitate the recovery of normal-light images with improved clarity and reduced noise from degraded low-light images. Utilizing this observation, RawNeRF \cite{mildenhall2022nerf} learns normally exposed scene representations from multi-view low-light raw images and renders novel views by incorporating a known camera internal ISP. However, as shown in Fig. \ref{Fig.1}, its effectiveness is substantially constrained by its reliance on the camera's pre-calibrated internal ISP, since calibrating the ISP of a specific camera is a tedious and complex process. Using a generic ISP offers a compromise, but this approach fails to achieve high-quality results across various cameras and scenarios, leading to suboptimal presentation and limited applicability.

In this paper, we propose a novel unsupervised method called \textbf{Bright-NeRF}  that enables the rendering of normal-light novel views from a set of low-light raw images. Our method directly reconstructs an enhanced normal-light radiance field using only low-light images for supervision and avoids the need for a pre-calibrated ISP for post-processing. Specifically, inspired by physically-based imaging theory, we model the sensor's raw-RGB response to scene illumination through collaborative learning with implicit scene representation, effectively compensating for the adverse effects of weak environmental illumination on image quality. Concurrently, we achieve effective noise suppression by aggregating information from multi-views. Leveraging the linear characteristics of raw images, we implement automatic exposure adjustment for low-light images. To address the significant performance degradation that can occur when transferring models trained on synthetic datasets to real-world scenarios (a challenge known as the ``sim-to-real gap"), we have collected a \textbf{L}ow-light \textbf{M}ulti-view \textbf{RAW} dataset (\textbf{LMRAW}). Our contributions are as follows:
\begin{itemize}
\item We propose an unsupervised framework, \textbf{Bright-NeRF}, that enables novel view synthesis under normal lighting conditions from a set of low-light raw images. Our approach effectively achieves noise suppression, color distortion correction, and brightness enhancement without the supervision of normal-light images.
\item We develop a model to characterize the sensor’s raw-RGB responses to scene illumination based on physically-inspired imaging theory, which mitigates color distortion caused by low-light conditions.
\item We have collected an \textbf{LMRAW} dataset. Extensive experiments and comparisons with various 2D image enhancement methods and NeRF-based approaches demonstrate that our method achieves state-of-the-art performance and maintains multi-view consistency on this dataset.

\end{itemize}

\begin{figure*}[t]
\centering
\includegraphics[width=1.0\textwidth]{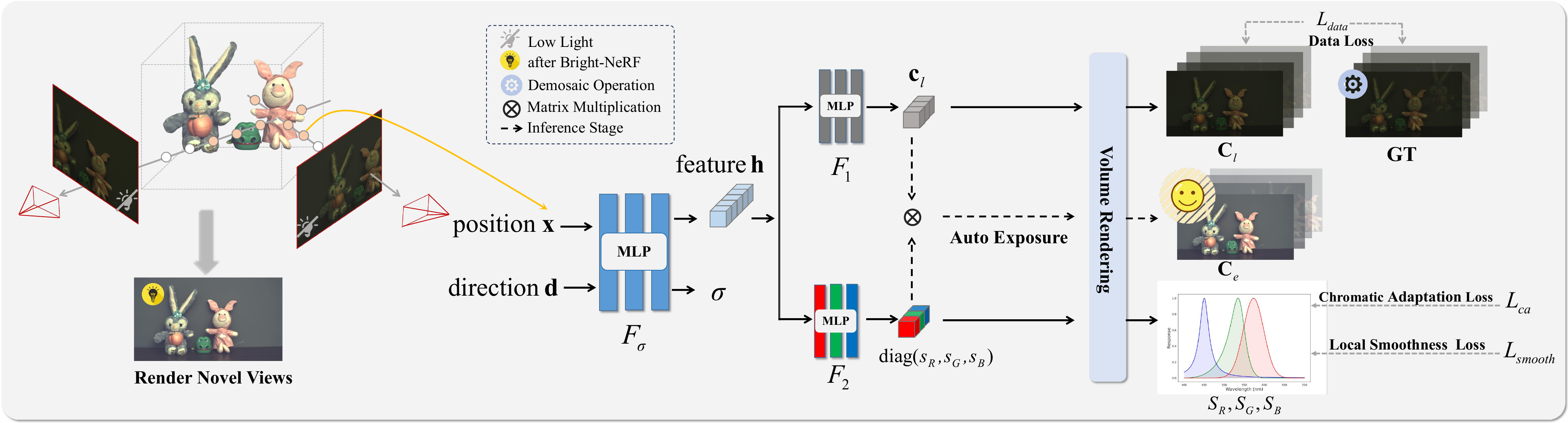} 
\caption{{\textbf{Method overview.}}
In the training stage, we firstly sample a series of points $\mathbf{x}$ from each ray and feed them into $F_\sigma$ along with the direction $\mathbf{d}$ to obtain the view-dependent features $\mathbf{h}$ and density $\sigma$ for each point. The low-light color $\mathbf{c}_l$ is then calculated by an MLP $F_1$. Another MLP $F_2$ is designed to estimate the sensor's adaptive response to low-photon lighting conditions $\text{diag}(s_R, s_G, s_B)$. The chromatic adaptation loss $L_{ca}$ optimizes the estimations regarding the sensor's response to address color distortion in low-light conditions, while the local smooth loss $L_{smooth}$ ensures smoothness in the learned response map $S_k, k\in{\{R, G, B\}}$. The data loss optimizes the modeling for low-light images.
In the inference stage, the estimated sensor's adaptive response $\text{diag}(s_R, s_G, s_B)$ is further applied to $\mathbf{c}_l$ to obtain distortion-corrected color and automatic exposure adjustment is achieved by utilizing the linear characteristics of raw data to increase the brightness. Through the joint learning of the low-light color field and the color restoration field, along with implicit noise smoothing and automatic exposure adjustment, a bright NeRF is eventually obtained.} 
\label{Fig.2}
\end{figure*}

\section{Related Work}

\subsection{Neural Radiance Field and Extensions}
NeRF \citep{mildenhall2021nerf} has drawn widespread attention due to its powerful representation of a scene using a set of posed images as input, they optimize a scene function via rendering the pixel color by volume rendering scheme along the corresponding ray, thus enabling novel views synthesis.

A primary limitation of NeRF lies in its assumption of ideal input views, which are captured under normal light without large illumination variations and do not contain blur. To address this constraint, researchers have proposed various innovative methods tailored to different types of scene degradation. Deblur-NeRF \citep{ma2022deblur} introduces a deformable sparse kernel module to explicitly simulate the physical blurring process. By synthesizing blurred images to match the input, it enables the recovery of clear scene representations from blurry images. NeRF-W \cite{martin2021nerf} focuses on modeling in-the-wild images characterized by illumination variations and transient objects by incorporating appearance embedding and transient embedding, implicitly integrating illumination changes and transient objects into the neural implicit representation. There is another line of work exploring the use of NeRF in low-light conditions. RawNeRF \cite{mildenhall2022nerf} proposes to train NeRF in the raw data domain and employ Image Signal Processing (ISP) to post-process the rendered images. LLNeRF \citep{wang2023lighting} decomposes the color of sampling points into lighting-related and reciprocal components, enabling the recovery of normal-light images from a set of low-light sRGB images. 
Despite the advancements in LLIE NeRF methods, they either depend on a pre-calibrated ISP for post-processing the renderings or struggle to recover accurate color details under extremely low-light conditions.

\subsection{Low-light Image Enhancement}
Low-light image enhancement aims to improve the visibility of images taken in insufficient illumination conditions while simultaneously suppressing noise, correcting color distortions, and preserving details. Deep learning-based image enhancement methods have demonstrated significant efficacy in this domain.  
LIME \cite{guo2016lime} proposes imposing structural priors to optimizer the illumination map, enabling to increase the naturalness of the enhancement.
Retinex-Net \cite{wei2018deep} introduces a dual-module structure comprising a reflectance module and an illumination adjustment module, combined with BM3D \cite{dabov2006image} for noise suppression. 

Recently some approaches have been proposed to directly retouch raw data from camera sensors into high-quality output images. SID \cite{chen2018learning} pioneers this research direction by proposing a fully convolutional neural network directly training on raw images. 
SGN \cite{gu2019self} adopts a top-down self-supervised structure to effectively exploit multi-scale image information. \cite{lamba2021restoring} proposes an amplifier module capable of directly estimating the amplification factor from input images, eliminating the need for ground truth exposure values to estimate the pre-amplification factor. DNF \cite{jin2023dnf} leverages the characteristics of raw and sRGB data domains to decouple domain-specific subtasks, avoiding domain ambiguity issues between raw and sRGB domains. 
However, these 2D methods are prone to overfitting the training images processed by a fixed ISP, leading to poor generalization across different cameras and scenarios.

\section{Preliminaries}
\subsection{Neural Radiance Field}
NeRF \cite{mildenhall2021nerf} represents a scene by learning a continuous function in 3D space, mapping a ray origin $\mathbf{o}$ and view direction $\mathbf{d}$ into volume density $\sigma$ and color $\mathbf{c}$. This process is defined as:
\begin{equation}
\begin{cases}
    (\mathbf{h}, \sigma) =  F_{\sigma}(\mathbf{o}, \mathbf{d}; \Theta_{F_{\sigma}}),  \;\\
    \mathbf{c} =  F_{1}(\mathbf{h}; \Theta_{F_{1}}),
\end{cases}
\label{Eq.1}
\end{equation}
where $\mathbf{h}$ is the view-dependent features learned by the neural network $F_{\sigma}$ and $F_1$ is the color field to output the view-dependent color.

More specifically, suppose a camera ray $\mathbf{r}$ is emitted from camera center $\mathbf{o}$ with direction $\mathbf{d}$, \textit{i.e.}, $\mathbf{r}(t) = \mathbf{o} + t\mathbf{d}$ where $t$ denotes the distance from the origin $\mathbf{o}$. The expected color $\mathbf{C}(\mathbf{r})$ of $\mathbf{r}$ is defined as:
\begin{equation}
  \mathbf{C}(\mathbf{r}) = \int_{t_{n}}^{t_{f}}T(t)\sigma(\mathbf{r}(t))\mathbf{c}(\mathbf{r}(t), \mathbf{d})dt,
 \label{Eq.2}
\end{equation}
where $t_{n}$ and $t_{f}$ denote the near and far boundary of the ray respectively.

\section{Method}
In this section, we introduce our method Bright-NeRF for recovering enhanced neural radiance fields. The overall pipeline of Bright-NeRF is summarized in Fig. \ref{Fig.2}. Our method aims to address the challenges inherent in low-light scenes, focusing on novel view synthesis with color restoration, noise reduction,  and brightness enhancement. We begin by detailing the architecture of Bright-NeRF, followed by a discussion on how we tackle these specific challenges. Finally, we outline our optimization strategies to improve the performance and quality of our method.

\subsection{Overall Framework}
Given a set of raw images captured under low-light conditions, our objective is to generate novel views as they would appear under normal lighting conditions. 
We assume that the density field of the scene representation under low-light and normal light conditions remains constant while the color field differs due to the influence of scene illumination. Specifically, the color field in low-light conditions exhibits lower brightness, stronger noise, and distorted color compared to well-lit conditions. Inspired by this observation, we propose learning a color restoration field from the view-dependent features $\mathbf{h}$, aiming to recover the adaptive sensor's response to low-photon conditions in the scene. This approach enables the rendering of vivid colors, noise-free, and properly illuminated scene representations.

Specifically, 
the low-light pixel values ${\mathbf{C}_{l}}(\mathbf{r})$ and the enhanced pixel values $\mathbf{C}_{e}(\mathbf{r})$ can be computed as: 
\begin{equation}
 \mathbf{C}_{l}(\mathbf{r}) = \int_{t_{n}}^{t_{f}}T(t)\sigma(\mathbf{r}(t)){\mathbf{c}}_{l}(\mathbf{r}(t), \mathbf{d})dt,
\label{Eq.4}
\end{equation}
\begin{equation}
 \mathbf{C}_{e}(\mathbf{r}) = \int_{t_{n}}^{t_{f}}T(t)\sigma(\mathbf{r}(t)){\mathbf{c}}_{e}(\mathbf{r}(t), \mathbf{d})dt,
 \label{Eq.5}
\end{equation}
where $\mathbf{{c}}_l$ and $\mathbf{{{c}}}_e$ represent the color at point $\mathbf{r}(t)$ along the ray $\mathbf{r}$ before and after enhancement, respectively. We then express the relationship between $\mathbf{{c}}_l$ and $\mathbf{{c}}_e$ as:
\begin{equation}
\mathbf{{c}}_{e} = \phi(\mathbf{{c}}_{l}),
\label{Eq.6}
\end{equation}
where $\phi$ is a comprehensive enhancement function capable of performing three key operations: denoising, color restoration, and brightness improvement. By decomposing the enhancement of the scene representation into these subtasks, our model can address each challenge more effectively in an unsupervised manner.

\subsection{Unsupervised Enhancement}

\subsubsection{Color Restoration.}
In low-light conditions, recovering an object's true color function is a complex task, as the reflected color depends on the object's intrinsic color and environmental illumination. Mathematically, to form a color image, consider a spatial point $\mathbf{x}$ along a ray $\mathbf{r}$ corresponding to a specific pixel, the color at this point can be expressed as:
\begin{equation}
    c_{k}(\mathbf{x}) = \int_{\omega}^{} c(\mathbf{x}, \lambda) s_k(\lambda) d\lambda,
\label{Eq.10}
\end{equation}
where $c(\mathbf{x}, \lambda)$ is the measurement at position $\mathbf{x}$ with respect to the wavelength $\lambda$, $s_{k}(\lambda)$ denotes the sensor characteristics as a function of wavelength $\lambda$, over the visible spectrum $\omega$. The subscript $k$ indicates the sensor's response in the $k$ channel ($k\in{\{R,G,B\}}$).
In low-light conditions, the unknown sensor characteristics of the low-photon environmental lighting make it challenging to obtain an illumination-invariant image descriptor. To address this issue, our model implicitly estimates the sensor's response to a low-light environment to eliminate the color distortion, enabling us to render the scene's tone independently of its unsatisfactory environmental illumination.

We adopt the von Kries adaptation hypothesis \cite{Brainard1992AsymmetricCM}, which suggests that the three color sensors in the human visual system each operate with independent gain control. Based on this principle, we implicitly estimate a diagonal matrix $\text{diag}(s_R, s_G, s_B)$ as follows:
\begin{equation}
\text{diag}({s}_R, {s}_G, {s}_B) = {F}_{2}(\mathbf{h}),    
\label{Eq.11}
\end{equation}
where $\mathbf{h}$ is the view-dependent feature derived by Eq. \ref{Eq.1}. ${F}_2$ is an MLP used to model the adaptive response of the visual system to low-photon lighting conditions. 
By applying the estimated sensor response to the predicted color $\mathbf{c}_l(\mathbf{x}, \mathbf{d})$ of a spatial point $\mathbf{x}$, we can recover the original color:
\begin{equation}
\mathbf{c}_s(\mathbf{x}, \mathbf{d}) = \mathbf{c}_l(\mathbf{x}, \mathbf{d}) \times \text{diag}({s}_R, {s}_G, {s}_B).
\label{Eq.12}
\end{equation}

Note that our estimated sensor's response can adaptively learn to accurately correct colors regardless of how dark the scene is, which is further analyzed in the experimental section.

\subsubsection{Brightness Improvement.}
There is a linear relationship exists between input brightness and output pixel values in raw signals:
\begin{equation}
    \mathbf{c}_{e}(\mathbf{r}) = \alpha \cdot \mathbf{c}_s(\mathbf{r}),
\label{Eq.13}
\end{equation}
where $\alpha$ is the proportionality coefficient.
Inspired by automatic exposure techniques, we dynamically adjust the transition curve slope $\alpha$ based on the average scene intensities to adapt to brightness levels.

\subsubsection{Denoising.}
We empirically adopt a heteroscedastic Gaussian process model in raw data: \citep{mohsen1975noise,liu2014practical}:
\begin{equation}
n(\mathbf{r}) \sim N(0, \beta^{2} \mathbf{c}_e(\mathbf{r}) + \delta^{2}),
\label{Eq.7}
\end{equation} 
where $\beta$ is a sensor-specific scaling factor of the signal and $\delta^{2}$ is the variance of the distribution.

Consider a spatial point $\mathbf{x}$ with a large density in the scene, which has multiple projections $P_{\mathbf{x}} = \left\{\widetilde{\mathbf{c}}_l(\mathbf{r})\right\}$ in the training images. For a color $\widetilde{\mathbf{c}}_l(\mathbf{r})$ and its underlying noise-free color $\bar{\mathbf{c}}_e(\mathbf{r})$,  their relationship can be written as:
\begin{equation}
    \widetilde{\mathbf{c}}_l(\mathbf{r}) = \bar{\mathbf{c}}_e(\mathbf{r}) + n(\mathbf{r}),
\label{Eq.8}
\end{equation}
where $n(\mathbf{r})$ represents the noise corrupting $\bar{\mathbf{c}}_e(\mathbf{r})$.
During training, the predicted color $\mathbf{c}_e(\mathbf{r})$ is supervised by all pixels in $P_{\mathbf{x}}$. Since the loss function for all rays corresponding to pixels in $P_{\mathbf{x}}$ is an unweighted average, the neural network tends to learn parameters that minimize the overall average deviation. Consequently, the learned $\mathbf{c}_e(\mathbf{r})$ would converge towards the expectation of $\widetilde{\mathbf{c}}_l(\mathbf{r})$, \textit{i.e.}, 
\begin{equation}
\mathbf{c}_e(\mathbf{r}) \approx \mathbb{E}\left\{\widetilde{\mathbf{c}}_l(\mathbf{r})\right\} = \bar{\mathbf{c}}_e(\mathbf{r}) + \mathbb{E}\{n\}.
\label{Eq.9}
\end{equation}

As the noise in raw signal is zero-mean, \textit{i.e.}, $\mathbb{E}\{n\} = 0$, it indicates that the multi-view optimization inherent in neural radiance fields can effectively smooth images and reduce the noise.

\begin{figure*}[t]
\centering
\includegraphics[width=1.0\textwidth]{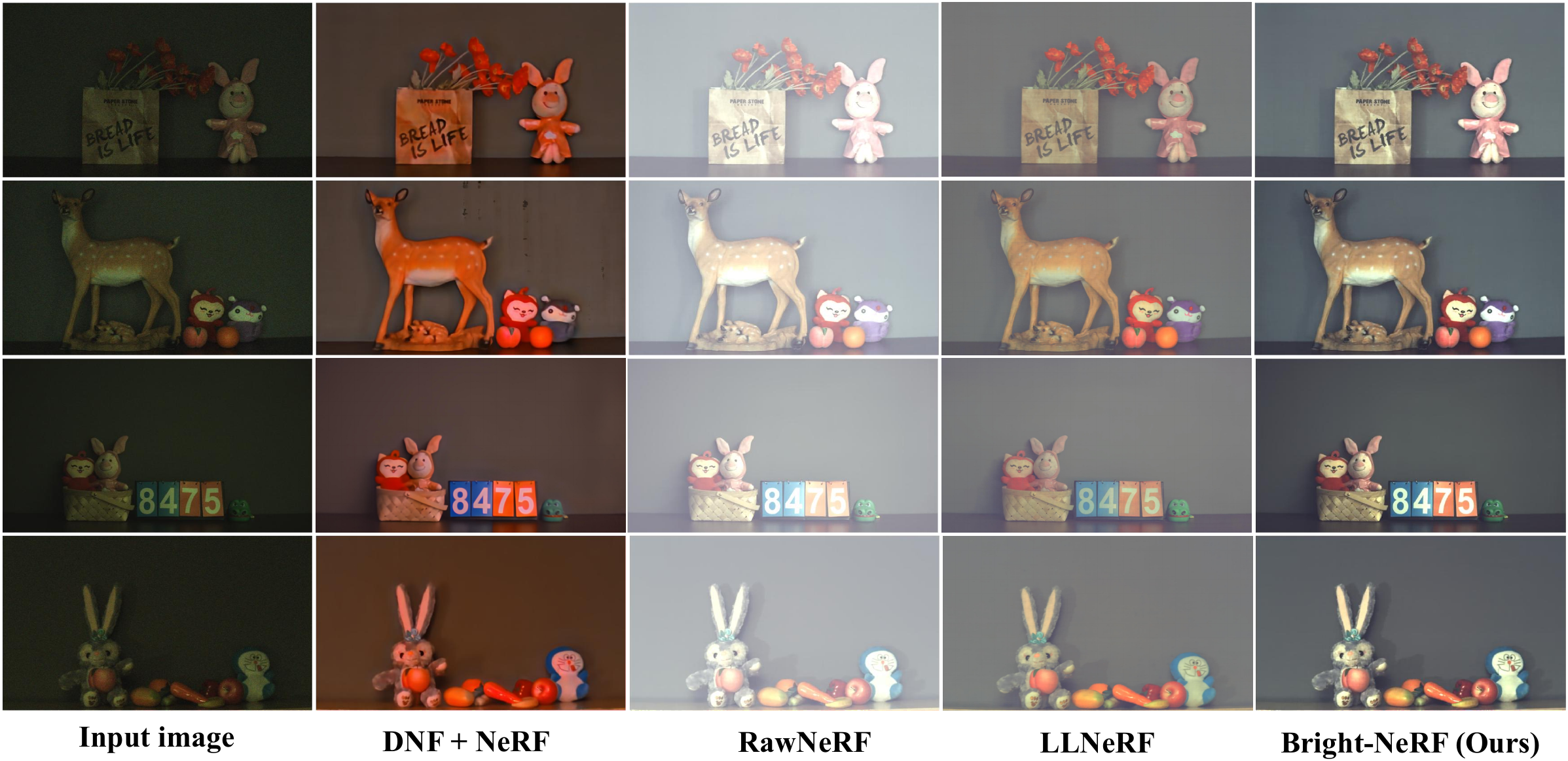} 
\caption{{\textbf{Qualitative comparison with state-of-the-art LLIE NeRF methods.}} Leveraging the sensor's adaptive responses in the color restoration field, the results of our Bright-NeRF exhibit more natural color and more accurate tone restoration.}
\label{Fig.3}
\end{figure*}

\subsection{Optimization}

In this section, we present the loss functions that guide the unsupervised training of Bright-NeRF: chromatic adaptation loss directs the optimization of the enhancement process, ensuring that the color adaptation is performed effectively and consistently in low-light conditions, and data loss is responsible for optimizing the radiance field. The combination of these loss functions enables Bright-NeRF to learn effectively without the need for paired low-light and normal-light image data, which is often difficult or expensive to obtain.

\subsubsection{Chromatic Adaptation Loss.}
To enhance the accuracy of our estimated sensor's response to low photon illumination in the scene, we introduce a chromatic adaptation loss, designed to constrain the learning of $\text{diag}({s}_R, {s}_G, {s}_B)$, promoting to produce more realistic images. The chromatic adaptation loss, denoted as $L_{ca}$, is formulated as:
\begin{equation}
   L_{ca} = \frac{1}{3} \sum \limits_{\mathbf{r} \in \mathbb{R}} \sum\limits_{k \in \{R, G, B\}} \left\Vert \frac{K_{avg}}{\bar{C_k}} - S_k(\mathbf{r}) \right\Vert^2,
\label{Eq.14}
\end{equation}
where $K_{avg}\!\!=\!\!\frac{1}{3}(\bar{C}_R\!+\!\bar{C}_G\!+\!\bar{C}_B)$ 
and $\bar{C}_k\!=\! \frac{1}{N} \sum \limits_{\mathbf{r} \in \mathbb{R}} C_k(\mathbf{r})$, $N$ is the number of sampled rays. 
And $S_k(\mathbf{r}) = \int_{t_{n}}^{t_{f}}T(t)\sigma(\mathbf{r}(t))s_k(\mathbf{r}(t), \mathbf{d})dt$, which denotes the integrated sensor's response to the low-light scene along the ray $\mathbf{r}$ for channel $k$.

\begin{figure*}[t]
\centering
\includegraphics[width=1.0\textwidth]{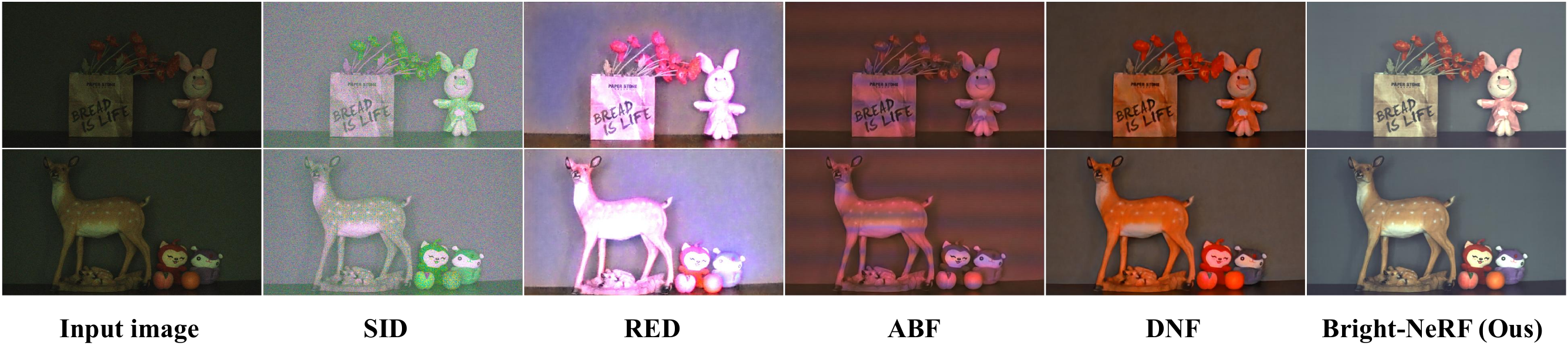} 
\caption{{\textbf{Qualitative comparison with state-of-the-art 2D LLIE methods.}} We show that our method effectively recovers detailed geometry and accurate colors while other methods often struggle with denoising and tend to produce incorrect colors. For these 2D methods, we use publicly available code and pre-trained model weights.}
\label{Fig.4}
\end{figure*}

\begin{table*}[h]

\centering

\begin{tabular}{ l|c|cccc } 

 \hline
 \hline
   Method & Method Type &  PSNR $\uparrow$ & SSIM $\uparrow$ & LPIPS $\downarrow$ & LIQE 
   $\uparrow $\\
 \hline
  SID \cite{chen2018learning} & \multirow{4}{*}{2D} & 13.37 & 0.233 & 0.786 & 1.004 
  \\
  RED \cite{lamba2021restoring} & & 11.04 & 0.654 & 0.731 & 1.003 
  \\
  ABF \cite{dong2022abandoning} & & 18.20 & 0.826 & 0.675 & 1.010 
  \\
  DNF \cite{jin2023dnf}& & 18.60 & 0.841 & 0.636 & 1.211 
  \\
  \hline
  RawNeRF \cite{mildenhall2022nerf} & \multirow{3}{*}{3D}& 8.95 & 0.694 & 0.524 & 1.626 
  \\
  DNF \cite{jin2023dnf}+NeRF & &18.20 & 0.829 & 0.663 & 1.151 
  \\
   
  LLNeRF \cite{wang2023lighting} & & 16.64 & 0.811 & 0.494 & 1.332
  \\
  \hline
  Bright-NeRF (Ours) & \multirow{1}{*}{3D} & \textbf{22.64} & \textbf{0.874} & \textbf{0.476} & \textbf{1.659} 
  \\

\hline
\hline 
\end{tabular}

\caption{\textbf{Quantitative comparison with existing LLIE methods.} The calculated metrics are the average values across nine scenes. The best results are marked in {\bf bold}.} 
 \label{Tab.1}
\end{table*}

\subsubsection{Local Smoothness Loss.}
To ensure that the estimated sensor‘s response to illumination maintains piece-wise smoothness, we impose the following constraint:
\begin{equation}
    L_{smooth} = \sum \limits_{\mathbf{r} \in \mathbb{R}} \left\Vert\frac{\gamma_1 * \mathbf{S}_v^2(\mathbf{r})}{\mathbf{C}_{h}^2(\mathbf{r}) + \epsilon} \right\Vert + \left\Vert \frac{\gamma_2 * \mathbf{S}_h^2(\mathbf{r})}{\mathbf{C}_{v}^2(\mathbf{r}) + \epsilon} \right\Vert,
\label{Eq.15}
\end{equation}
where $\mathbf{S}_v$, $\mathbf{S}_h$ and $\mathbf{C}_v$, $\mathbf{C}_h$ represent the differences in the integral of sensor responses and colors between adjacent rays in the vertical and horizontal directions, respectively. $\gamma_1$ and $\gamma_2$ are weighting factors and $\epsilon$ is a small positive constant to prevent division by zero and ensure numerical stability.
\subsubsection{Data Loss.}
Given that the pixels in our training data are predominantly low intensity due to the low-light conditions, we adopt the color loss function proposed in \cite{mildenhall2022nerf}, which 
amplifies the error in dark regions, making the optimization process more sensitive to subtle differences in low-light areas.
The data loss is formulated as:
\begin{equation}
L_{data} = \sum\limits_{\mathbf{r}\in \mathbb{R}}\Vert\psi(\mathbf{C}_{l}(\mathbf{r})) - \psi({\mathbf{C}}_{GT}(\mathbf{r}))\Vert^{2},    
\label{Eq.16}
\end{equation}
where ${\mathbf{C}}_l$ represents the rendered low-light color and
$\mathbf{C}_{GT}$ denotes the low-light ground truth color.
$\psi(y) = log(y+\epsilon^{\prime})$ is the linearized tone mapping function.

Above all, our loss function $L$ comprises three components: data loss $L_{data}$, along with two unsupervised losses $L_{ca}$ and $L_{smooth}$.
\begin{equation}
    L = \lambda_1 L_{data} + \lambda_2 L_{ca} + \lambda_3 L_{smooth},
\label{Eq.17}
\end{equation}
where $\lambda_1$, $\lambda_2$, and $\lambda_3$ are three positive weights, which are set to 1.0, 0.1, and 0.1 respectively.

\subsubsection{Train and Test Schemes.}
At the training phase, Bright-NeRF simultaneously renders low-light images ${\mathbf{C}}_l$ and incorporates the estimated sensor's response $\text{diag}({s}_R, {s}_G, {s}_B)$ to render the sensor's response map $\mathbf{S}$ at the volume rendering stage. At the testing stage, we leverage the comprehensive enhancement function in Eq. \ref{Eq.6} and perform volume rendering in Eq. \ref{Eq.5} to directly produce enhanced normal-light images ${\mathbf{C}}_e$.

\section{Experiments}

\subsection{Dataset}
To facilitate the training and evaluation of our model, we collect a multi-view low-light raw image dataset LMRAW. While previous work \cite{mildenhall2022nerf} includes some low-light scenarios, their focus is primarily on raw denoising rather than low-light enhancement. The absence of ground truth for low-light conditions in their test scenes makes it difficult to assess the enhancement performance for novel view synthesis.

Our dataset comprises scenes captured under challenging low-light conditions with illumination levels ranging from 0.01 to 0.05 lux. Each scene consists of 18 to 30 images at $1920 \times 1080$ resolution. To capture multi-view images, we move and rotate the camera tripod. Since COLMAP \cite{schonberger2016structure} does not support processing raw images directly, we use full-resolution post-processed JPEG images for pose estimation.

\subsection{Results}

\subsubsection{Novel View Synthesis.}
To ensure a fair comparison, we train our Bright-NeRF, the baseline model, RawNeRF \cite{mildenhall2022nerf} using the same images, and train LLNeRF \cite{wang2023lighting} using post-processed RGB data as it is a RGB-based method. The comparison of novel view synthesis results is illustrated in Fig. \ref{Fig.3}. For the baseline, we choose to pre-enhance the low-light training images using DNF \cite{jin2023dnf} as it provides better geometric information and subsequently trains a NeRF model with these enhanced images. We can see that the baseline model’s results show noticeable discrepancies in color reproduction compared to the expected natural appearance of the scenes.
RawNeRF \cite{mildenhall2022nerf} struggles with color distortion in extremely dark scenes without a calibrated ISP which is a intricate process, and a generic ISP yields suboptimal performance. Their results exhibit a general washout effect, resulting in hazy appearance and unnatural color reproduction.
The rendered results produced by LLNeRF \cite{wang2023lighting} demonstrate inadequate restoration of tone and saturation. In contrast, our model exhibits superior performance in producing more natural and vivid colors thanks to our color restoration strategy .

\begin{figure}[h]
\centering
\includegraphics[width=1\columnwidth]{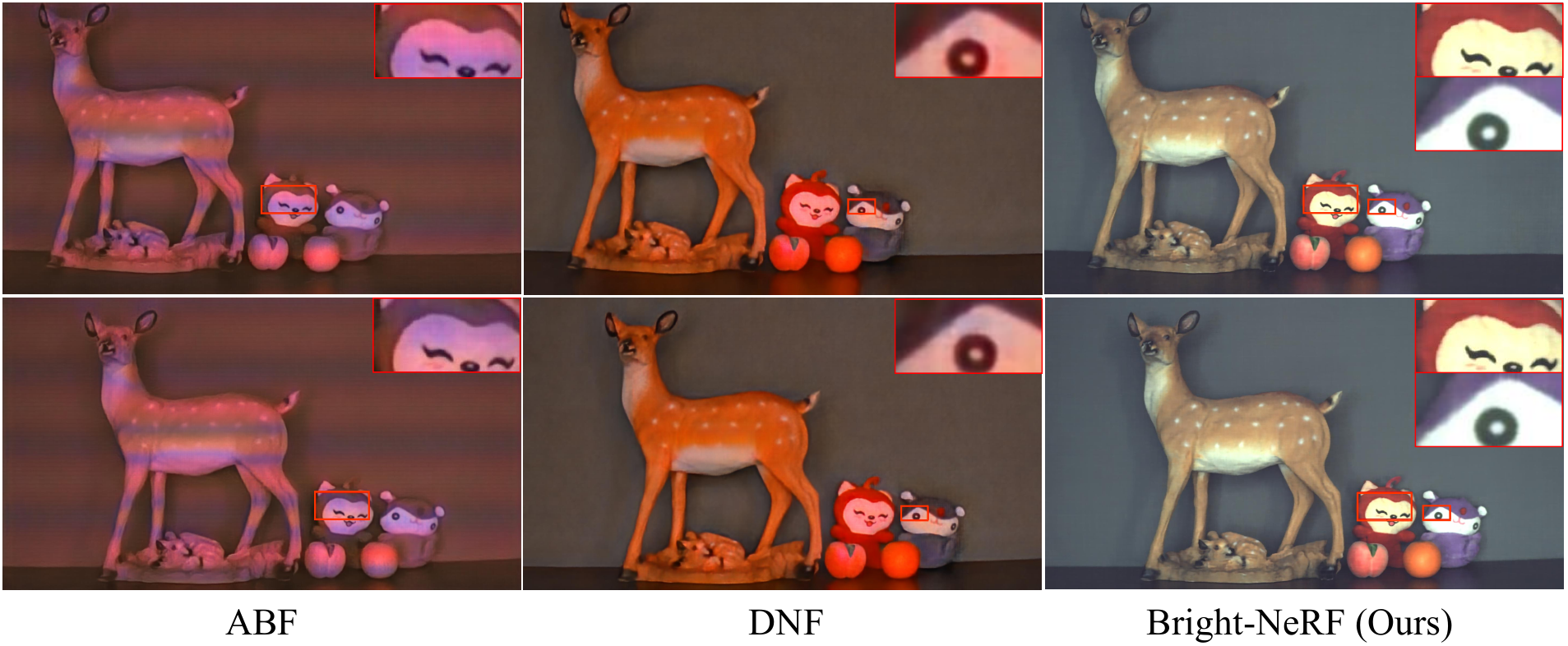} 
\caption{{\textbf{Visualization of multi-view consistency in 2D LLIE methods and our method.}} We present cropped regions of enhancement results from the ABF \cite{dong2022abandoning} and DNF \cite{jin2023dnf} methods at different views. Both methods exhibit enhancement inconsistency such as the color inconsistency in the doll's ``forehead" in the first column and ``eye" in the second column. In contrast, our Bright-NeRF ensures high-quality enhancement while maintaining multi-view consistency. }
\label{Fig.5}
\end{figure}

Beyond analyzing the performance of the LLIE NeRF methods, we further investigate the consistency of 2D image enhancement methods during the multi-view enhancement process. As illustrated in Fig. \ref{Fig.5}, when images from two different views are input into ABF \cite{dong2022abandoning} and DNF \cite{jin2023dnf}, they tend to produce inconsistent artifacts as they enhance each view independently and do not consider the consistency between views. Compared to 2D enhancement methods, our approach guarantee the multi-view consistency in enhancement, ensuring high-quality rendering results.

\subsubsection{Low-light Enhancement.}
We conducted experimental comparisons with state-of-the-art 2D raw image low-light enhancement methods, as shown in Fig. \ref{Fig.4}. We use the publicly available code and pre-trained model weights to enhance the images. Since SID \cite{chen2018learning} and RED \cite{lamba2021restoring} are trained on the dataset composed of relatively brighter images, they struggle to handle significant noise in darker scenes. While ABF \cite{dong2022abandoning} and DNF \cite{jin2023dnf} can reduce the noise, they tend to overfit to the data processed by the ISP used for supervision in the training dataset, failing to recover accurate and natural colors in diverse camera and scenarios. In contrast, our model effectively improves image brightness while simultaneously reducing noise, recovering natural colors from the color distortions present in low-light conditions.

\subsubsection{Adaptability and Stability across Different Darkness
Degrees.}
\begin{figure}[h]
    \centering
    \includegraphics[width=1\linewidth]{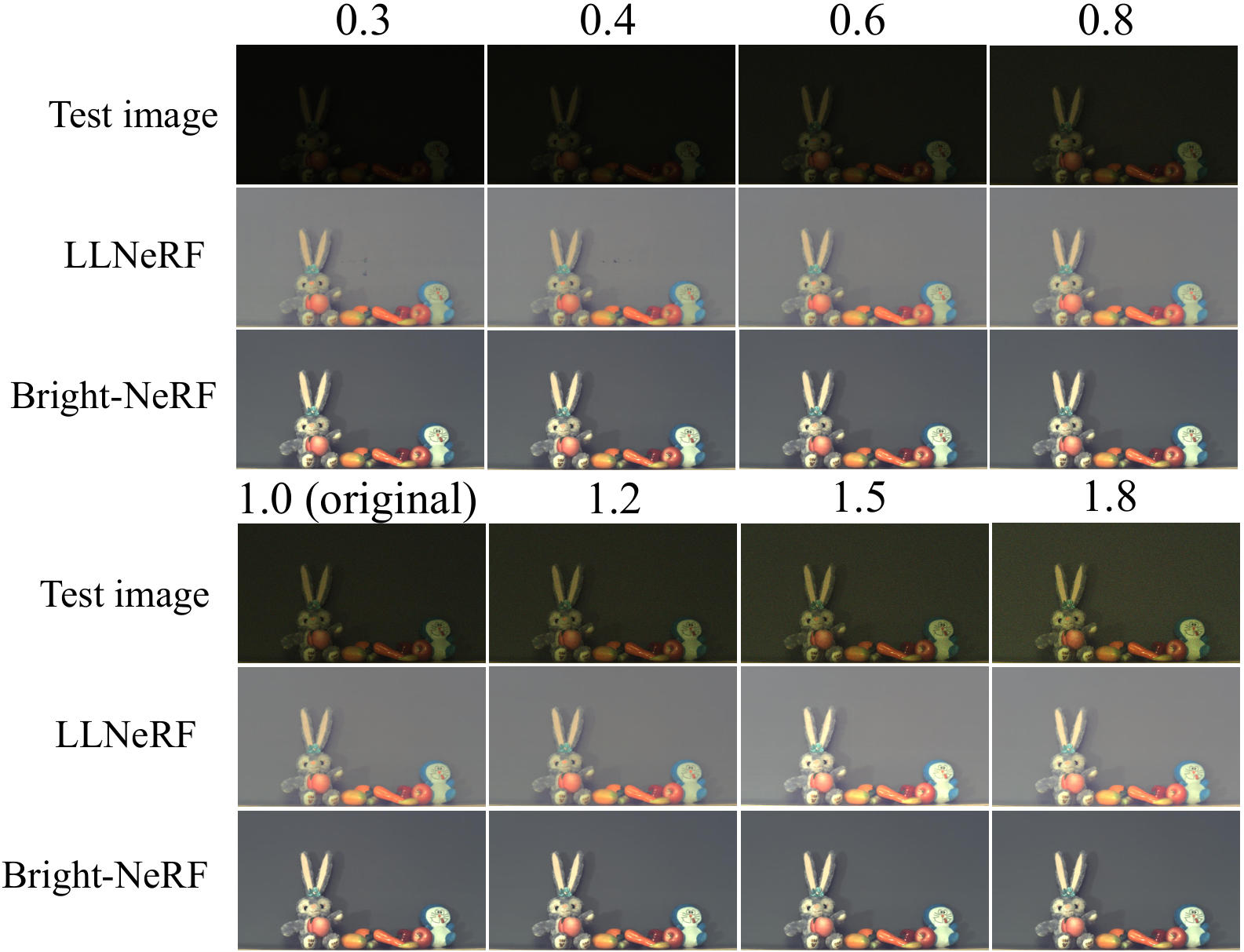}
    \caption{
    \textbf{Performance comparison on scenes with different darkness degrees.}
    As the simulated exposure ratio decreases (darker scene), LLNeRF's results exhibit more severe color bias while our Bright-NeRF demonstrates adaptability and stability to different degrees of darkness, producing natural and stable color even in extremely low-light conditions. The simulated exposure ratio is defined as the ratio of the exposure time compared to the original raw image.}
    \label{Fig.6}
\end{figure}

To investigate the adaptability and stability of our method across different brightness levels, we simulate scenes with different darkness degrees. The simulated exposure ratio is defined as the ratio
of the exposure time compared to the original raw image. Leveraging the linear characteristics of raw data \cite{zhu2020eemefn}, we generate 8 sets of scene darkness by simulating the exposure ratio and make a comparison with LLNeRF \cite{wang2023lighting} which achieves the second best performance in existing LLIE NeRF methods, as shown in Fig. \ref{Fig.6}.  
As the scene becomes darker, LLNeRF suffers from increasingly severe color bias. 
We attribute this to the difficulty LLNeRF faces in accurately decomposing colors into lighting-related and reciprocal components in extremely dark scenes, as well as the limitations of low bit-depth RGB data in recovering color details. Despite these challenges, our method makes use of the high-bit characteristic of raw data and uses the color restoration field to produce accurate color details, even under extremely low exposure ratios such as 0.3 and 0.4. Our method adaptively renders results with stable brightness and natural color regardless of the scene's darkness degree. We refer to the implementation of simulated exposure and more comparisons in the supplementary material. 

\subsubsection{Quantitative Evaluation.}
We evaluate our model's performance across nine scenes, as shown in Tab. \ref{Tab.1}. Our method outperforms existing 2D and LLIE NeRF approaches across PSNR, SSIM, and LPIPS, which indicates that our method performs better in relative sharpness, structure, and perceptual quality. 

To further evaluate the rendering and perceptual quality of our method, we leverage the latest non-reference image quality assessment metrics, LIQE \cite{zhang2023blind}, which integrates multi-modality into low-level vision perception, ensuring robustness in image quality assessment across diverse scenarios. The comparison on the LIQE metric indicates that the proposed method delivers superior performance in terms of perceptual image quality without any reference information.
\subsubsection{Ablation Study.}

As shown in Tab. \ref{Tab.2}, we analyze the impact of the color restoration field by removing the branch responsible for estimating the sensor's response to scene illumination and its related chromatic adaptation loss.
This results in the color field being affected by color distortion, leading to incorrect rendering and performance degradation. With the proposed adaptive color restoration field, we achieve accurate perception of environmental low-photon conditions, enabling correct color rendering. Additionally, by relaxing the constraints of the loss functions, we observe a decline in the quality of results when any item is removed.

\begin{table}[h]
  \centering

    \begin{tabularx}{\linewidth}{cc|XXXX}
    \hline    
    \hline    
    ${CR} + {L_{ca}}$  & ${L_{smooth}}$ & PSNR$\uparrow$ & SSIM$\uparrow$ & LPIPS$\downarrow$ & LIQE$\uparrow$\\
    \hline    
       \xmark &  \xmark  & 21.37 & 0.868 & 0.737 & 1.497 \\
        \cmark  & \xmark  & 22.61 & 0.873 & 0.478 & 1.614 \\
        \cmark & \cmark  & \textbf{22.64} & \textbf{0.874} & \textbf{0.476} & \textbf{1.659} \\
    \hline    
    \hline    
    \end{tabularx}%

\caption{\textbf{Ablation study results.} The quality of results is degraded as we remove any item. $CR$ denotes ``Color Restoration".} 
\label{Tab.2}
\end{table}

\section{Conclusion}
In this paper, we have proposed a novel unsupervised method for synthesizing novel views from a set of low-light raw images, simultaneously addressing the challenges of denoising, color restoration, and brightness improvement. Specifically, our method estimates the sensor's response to low-light scenes jointly with the scene's radiance field, enabling vivid color restoration and effective noise reduction regardless of the scene’s darkness degree. Our method innovatively enhances scene representation without relying on well-lit supervisions and achieves satisfactory rendering results. We have conducted extensive experimental evaluations to demonstrate its effectiveness in comparison to existing state-of-the-art approaches.

\section{Acknowledgments}
This work was supported by NSFC
under Grant 62031023.

\bibliography{aaai25}

\begin{thebibliography}{34}
\providecommand{\natexlab}[1]{#1}

\bibitem[{Adamkiewicz et~al.(2022)Adamkiewicz, Chen, Caccavale, Gardner,
  Culbertson, Bohg, and Schwager}]{adamkiewicz2022vision}
Adamkiewicz, M.; Chen, T.; Caccavale, A.; Gardner, R.; Culbertson, P.; Bohg,
  J.; and Schwager, M. 2022.
\newblock Vision-only robot navigation in a neural radiance world.
\newblock \emph{IEEE Robotics and Automation Letters}, 7(2): 4606--4613.

\bibitem[{Barron et~al.(2021)Barron, Mildenhall, Tancik, Hedman,
  Martin-Brualla, and Srinivasan}]{barron2021mip}
Barron, J.~T.; Mildenhall, B.; Tancik, M.; Hedman, P.; Martin-Brualla, R.; and
  Srinivasan, P.~P. 2021.
\newblock Mip-nerf: A multiscale representation for anti-aliasing neural
  radiance fields.
\newblock In \emph{Proceedings of the IEEE/CVF international conference on
  computer vision}, 5855--5864.

\bibitem[{Barron et~al.(2022)Barron, Mildenhall, Verbin, Srinivasan, and
  Hedman}]{barron2022mip}
Barron, J.~T.; Mildenhall, B.; Verbin, D.; Srinivasan, P.~P.; and Hedman, P.
  2022.
\newblock Mip-nerf 360: Unbounded anti-aliased neural radiance fields.
\newblock In \emph{Proceedings of the IEEE/CVF conference on computer vision
  and pattern recognition}, 5470--5479.

\bibitem[{Brainard and Wandell(1992)}]{Brainard1992AsymmetricCM}
Brainard, D.~H.; and Wandell, B.~A. 1992.
\newblock Asymmetric color matching: how color appearance depends on the
  illuminant.
\newblock \emph{Journal of the Optical Society of America. A, Optics and image
  science}, 9 9: 1433--48.

\bibitem[{Chen et~al.(2018)Chen, Chen, Xu, and Koltun}]{chen2018learning}
Chen, C.; Chen, Q.; Xu, J.; and Koltun, V. 2018.
\newblock Learning to see in the dark.
\newblock In \emph{Proceedings of the IEEE conference on computer vision and
  pattern recognition}, 3291--3300.

\bibitem[{Dabov et~al.(2006)Dabov, Foi, Katkovnik, and
  Egiazarian}]{dabov2006image}
Dabov, K.; Foi, A.; Katkovnik, V.; and Egiazarian, K. 2006.
\newblock Image denoising with block-matching and 3D filtering.
\newblock In \emph{Image processing: algorithms and systems, neural networks,
  and machine learning}, volume 6064, 354--365. SPIE.

\bibitem[{Dong et~al.(2022)Dong, Xu, Miao, Ma, Zhang, Yang, Jin, Teoh, and
  Shen}]{dong2022abandoning}
Dong, X.; Xu, W.; Miao, Z.; Ma, L.; Zhang, C.; Yang, J.; Jin, Z.; Teoh, A.
  B.~J.; and Shen, J. 2022.
\newblock Abandoning the bayer-filter to see in the dark.
\newblock In \emph{Proceedings of the IEEE/CVF conference on computer vision
  and pattern recognition}, 17431--17440.

\bibitem[{Gu et~al.(2019)Gu, Li, Gool, and Timofte}]{gu2019self}
Gu, S.; Li, Y.; Gool, L.~V.; and Timofte, R. 2019.
\newblock Self-guided network for fast image denoising.
\newblock In \emph{Proceedings of the IEEE/CVF International Conference on
  Computer Vision}, 2511--2520.

\bibitem[{Guo, Li, and Ling(2016)}]{guo2016lime}
Guo, X.; Li, Y.; and Ling, H. 2016.
\newblock LIME: Low-light image enhancement via illumination map estimation.
\newblock \emph{IEEE Transactions on image processing}, 26(2): 982--993.

\bibitem[{Hu et~al.(2023)Hu, Huang, Liu, Tai, and Tang}]{hu2023nerf}
Hu, B.; Huang, J.; Liu, Y.; Tai, Y.-W.; and Tang, C.-K. 2023.
\newblock Nerf-rpn: A general framework for object detection in nerfs.
\newblock In \emph{Proceedings of the IEEE/CVF Conference on Computer Vision
  and Pattern Recognition}, 23528--23538.

\bibitem[{Huang et~al.(2022)Huang, Zhang, Feng, Li, Wang, and
  Wang}]{huang2022hdr}
Huang, X.; Zhang, Q.; Feng, Y.; Li, H.; Wang, X.; and Wang, Q. 2022.
\newblock Hdr-nerf: High dynamic range neural radiance fields.
\newblock In \emph{Proceedings of the IEEE/CVF Conference on Computer Vision
  and Pattern Recognition}, 18398--18408.

\bibitem[{Jin et~al.(2023)Jin, Han, Li, Guo, Chai, and Li}]{jin2023dnf}
Jin, X.; Han, L.-H.; Li, Z.; Guo, C.-L.; Chai, Z.; and Li, C. 2023.
\newblock Dnf: Decouple and feedback network for seeing in the dark.
\newblock In \emph{Proceedings of the IEEE/CVF Conference on Computer Vision
  and Pattern Recognition}, 18135--18144.

\bibitem[{Kingma(2014)}]{kingma2014adam}
Kingma, D. 2014.
\newblock Adam: a method for stochastic optimization.
\newblock \emph{arXiv preprint arXiv:1412.6980}.

\bibitem[{Lamba and Mitra(2021)}]{lamba2021restoring}
Lamba, M.; and Mitra, K. 2021.
\newblock Restoring extremely dark images in real time.
\newblock In \emph{Proceedings of the IEEE/CVF conference on computer vision
  and pattern recognition}, 3487--3497.

\bibitem[{Laroche and Prescott(1993)}]{1993Apparatus}
Laroche, C.~A.; and Prescott, M.~A. 1993.
\newblock Apparatus and method for adaptively interpolating a full color image
  utilizing chrominance gradients.

\bibitem[{Lee et~al.(2023)Lee, Lee, Shin, and Lee}]{lee2023dp}
Lee, D.; Lee, M.; Shin, C.; and Lee, S. 2023.
\newblock Dp-nerf: Deblurred neural radiance field with physical scene priors.
\newblock In \emph{Proceedings of the IEEE/CVF Conference on Computer Vision
  and Pattern Recognition}, 12386--12396.

\bibitem[{Liu, Tanaka, and Okutomi(2014)}]{liu2014practical}
Liu, X.; Tanaka, M.; and Okutomi, M. 2014.
\newblock Practical signal-dependent noise parameter estimation from a single
  noisy image.
\newblock \emph{IEEE Transactions on Image Processing}, 23(10): 4361--4371.

\bibitem[{Ma et~al.(2022{\natexlab{a}})Ma, Li, Liao, Zhang, Wang, Wang, and
  Sander}]{ma2022deblur}
Ma, L.; Li, X.; Liao, J.; Zhang, Q.; Wang, X.; Wang, J.; and Sander, P.~V.
  2022{\natexlab{a}}.
\newblock Deblur-nerf: Neural radiance fields from blurry images.
\newblock In \emph{Proceedings of the IEEE/CVF Conference on Computer Vision
  and Pattern Recognition}, 12861--12870.

\bibitem[{Ma et~al.(2022{\natexlab{b}})Ma, Ma, Liu, Fan, and
  Luo}]{ma2022toward}
Ma, L.; Ma, T.; Liu, R.; Fan, X.; and Luo, Z. 2022{\natexlab{b}}.
\newblock Toward fast, flexible, and robust low-light image enhancement.
\newblock In \emph{Proceedings of the IEEE/CVF conference on computer vision
  and pattern recognition}, 5637--5646.

\bibitem[{Martin-Brualla et~al.(2021)Martin-Brualla, Radwan, Sajjadi, Barron,
  Dosovitskiy, and Duckworth}]{martin2021nerf}
Martin-Brualla, R.; Radwan, N.; Sajjadi, M.~S.; Barron, J.~T.; Dosovitskiy, A.;
  and Duckworth, D. 2021.
\newblock Nerf in the wild: Neural radiance fields for unconstrained photo
  collections.
\newblock In \emph{Proceedings of the IEEE/CVF conference on computer vision
  and pattern recognition}, 7210--7219.

\bibitem[{Mildenhall et~al.(2022)Mildenhall, Hedman, Martin-Brualla,
  Srinivasan, and Barron}]{mildenhall2022nerf}
Mildenhall, B.; Hedman, P.; Martin-Brualla, R.; Srinivasan, P.~P.; and Barron,
  J.~T. 2022.
\newblock Nerf in the dark: High dynamic range view synthesis from noisy raw
  images.
\newblock In \emph{Proceedings of the IEEE/CVF conference on computer vision
  and pattern recognition}, 16190--16199.

\bibitem[{Mildenhall et~al.(2021)Mildenhall, Srinivasan, Tancik, Barron,
  Ramamoorthi, and Ng}]{mildenhall2021nerf}
Mildenhall, B.; Srinivasan, P.~P.; Tancik, M.; Barron, J.~T.; Ramamoorthi, R.;
  and Ng, R. 2021.
\newblock Nerf: Representing scenes as neural radiance fields for view
  synthesis.
\newblock \emph{Communications of the ACM}, 65(1): 99--106.

\bibitem[{Mohsen, Tompsett, and S{\`e}quin(1975)}]{mohsen1975noise}
Mohsen, A.~M.; Tompsett, M.~F.; and S{\`e}quin, C.~H. 1975.
\newblock Noise measurements in charge-coupled devices.
\newblock \emph{IEEE Transactions on Electron Devices}, 22(5): 209--218.

\bibitem[{Rudnev et~al.(2022)Rudnev, Elgharib, Smith, Liu, Golyanik, and
  Theobalt}]{rudnev2022nerf}
Rudnev, V.; Elgharib, M.; Smith, W.; Liu, L.; Golyanik, V.; and Theobalt, C.
  2022.
\newblock Nerf for outdoor scene relighting.
\newblock In \emph{European Conference on Computer Vision}, 615--631. Springer.

\bibitem[{Schonberger and Frahm(2016)}]{schonberger2016structure}
Schonberger, J.~L.; and Frahm, J.-M. 2016.
\newblock Structure-from-motion revisited.
\newblock In \emph{Proceedings of the IEEE conference on computer vision and
  pattern recognition}, 4104--4113.

\bibitem[{Song et~al.(2023)Song, Choi, Do, Lee, and Kim}]{song2023blending}
Song, H.; Choi, S.; Do, H.; Lee, C.; and Kim, T. 2023.
\newblock Blending-nerf: Text-driven localized editing in neural radiance
  fields.
\newblock In \emph{Proceedings of the IEEE/CVF International Conference on
  Computer Vision}, 14383--14393.

\bibitem[{Tonderski et~al.(2024)Tonderski, Lindstr{\"o}m, Hess, Ljungbergh,
  Svensson, and Petersson}]{tonderski2024neurad}
Tonderski, A.; Lindstr{\"o}m, C.; Hess, G.; Ljungbergh, W.; Svensson, L.; and
  Petersson, C. 2024.
\newblock Neurad: Neural rendering for autonomous driving.
\newblock In \emph{Proceedings of the IEEE/CVF Conference on Computer Vision
  and Pattern Recognition}, 14895--14904.

\bibitem[{Wang et~al.(2023)Wang, Xu, Xu, and Lau}]{wang2023lighting}
Wang, H.; Xu, X.; Xu, K.; and Lau, R.~W. 2023.
\newblock Lighting up nerf via unsupervised decomposition and enhancement.
\newblock In \emph{Proceedings of the IEEE/CVF International Conference on
  Computer Vision}, 12632--12641.

\bibitem[{Wei et~al.(2018)Wei, Wang, Yang, and Liu}]{wei2018deep}
Wei, C.; Wang, W.; Yang, W.; and Liu, J. 2018.
\newblock Deep retinex decomposition for low-light enhancement.
\newblock \emph{arXiv preprint arXiv:1808.04560}.

\bibitem[{Xu et~al.(2023)Xu, Wu, Hou, Tsai, Li, Wang, Zhan, He, Vajda, Keutzer
  et~al.}]{xu2023nerf}
Xu, C.; Wu, B.; Hou, J.; Tsai, S.; Li, R.; Wang, J.; Zhan, W.; He, Z.; Vajda,
  P.; Keutzer, K.; et~al. 2023.
\newblock Nerf-det: Learning geometry-aware volumetric representation for
  multi-view 3d object detection.
\newblock In \emph{Proceedings of the IEEE/CVF International Conference on
  Computer Vision}, 23320--23330.

\bibitem[{Yuan et~al.(2022)Yuan, Sun, Lai, Ma, Jia, and Gao}]{yuan2022nerf}
Yuan, Y.-J.; Sun, Y.-T.; Lai, Y.-K.; Ma, Y.; Jia, R.; and Gao, L. 2022.
\newblock Nerf-editing: geometry editing of neural radiance fields.
\newblock In \emph{Proceedings of the IEEE/CVF Conference on Computer Vision
  and Pattern Recognition}, 18353--18364.

\bibitem[{Zhang et~al.(2023)Zhang, Zhai, Wei, Yang, and Ma}]{zhang2023blind}
Zhang, W.; Zhai, G.; Wei, Y.; Yang, X.; and Ma, K. 2023.
\newblock Blind image quality assessment via vision-language correspondence: A
  multitask learning perspective.
\newblock In \emph{Proceedings of the IEEE/CVF conference on computer vision
  and pattern recognition}, 14071--14081.

\bibitem[{Zhang et~al.(2021)Zhang, Srinivasan, Deng, Debevec, Freeman, and
  Barron}]{zhang2021nerfactor}
Zhang, X.; Srinivasan, P.~P.; Deng, B.; Debevec, P.; Freeman, W.~T.; and
  Barron, J.~T. 2021.
\newblock Nerfactor: Neural factorization of shape and reflectance under an
  unknown illumination.
\newblock \emph{ACM Transactions on Graphics (ToG)}, 40(6): 1--18.

\bibitem[{Zhu et~al.(2020)Zhu, Pan, Chen, and Yang}]{zhu2020eemefn}
Zhu, M.; Pan, P.; Chen, W.; and Yang, Y. 2020.
\newblock Eemefn: Low-light image enhancement via edge-enhanced multi-exposure
  fusion network.
\newblock In \emph{Proceedings of the AAAI conference on artificial
  intelligence}, volume~34, 13106--13113.

\end{thebibliography}

\clearpage
\appendix

\section{Implementation Details}
\subsubsection{Our Method}
Our code is built upon the Jax implementation of RawNeRF \cite{mildenhall2022nerf}. The training of our Bright-NeRF is conducted on a single NVIDIA GeForce RTX 3090 GPU with 24G of memory, running on the Ubuntu operating system. During training, we use full-resolution images and do not downsample the inputs to avoid affecting the noise distribution of raw linear data. We use the Adam optimizer \cite{kingma2014adam} and the optimization's learning rate decays from $10^{-3}$ to $10^{-5}$ over 105k steps. The batch size of rays is set to 1024.
\subsection{Baseline and Existing LLIE NeRF Methods}
To ensure a fair comparison, the important parameters in the baseline model and state-of-the-art LLIE NeRF models such as the number of samples per ray, batch size, and the number of training steps are set as same as these of us.
\subsubsection{Baseline}
We input the mosaiced raw images into the DNF \cite{jin2023dnf} to obtain enhanced RGB images, which are then used to train NeRF \cite{mildenhall2021nerf}.
\subsubsection{LLNeRF and RawNeRF}
Since LLNeRF \cite{wang2023lighting} is based on sRGB, we pre-process our raw dataset using open-source ISP processing steps to obtain sRGB data. The processing steps include: demosaicing \cite{1993Apparatus}, white balance in the OpenCV library function, and gamma correction. The post-processing steps applied to images rendered by RawNeRF \cite{mildenhall2022nerf} is identical to the ISP processing methods mentioned above. 

\section{Dataset}
Most existing multi-view datasets simulate low-light scenes by reducing exposure time and ISO, neglecting the impact posed by low photon counts and the significant noise introduced by high ISO. To create a more realistic setting, we consider environmental illumination and capture our dataset in a controlled dark-room environment, with illumination levels in the LMRAW dataset ranging from 0.01 lux to 0.05 lux. Fig. ~\ref{Fig.5} shows the example images of scenes from our proposed dataset.

\section{Simulated Exposure Ratio Discussion}
\begin{figure*}
\centering
\includegraphics[width=1\textwidth]{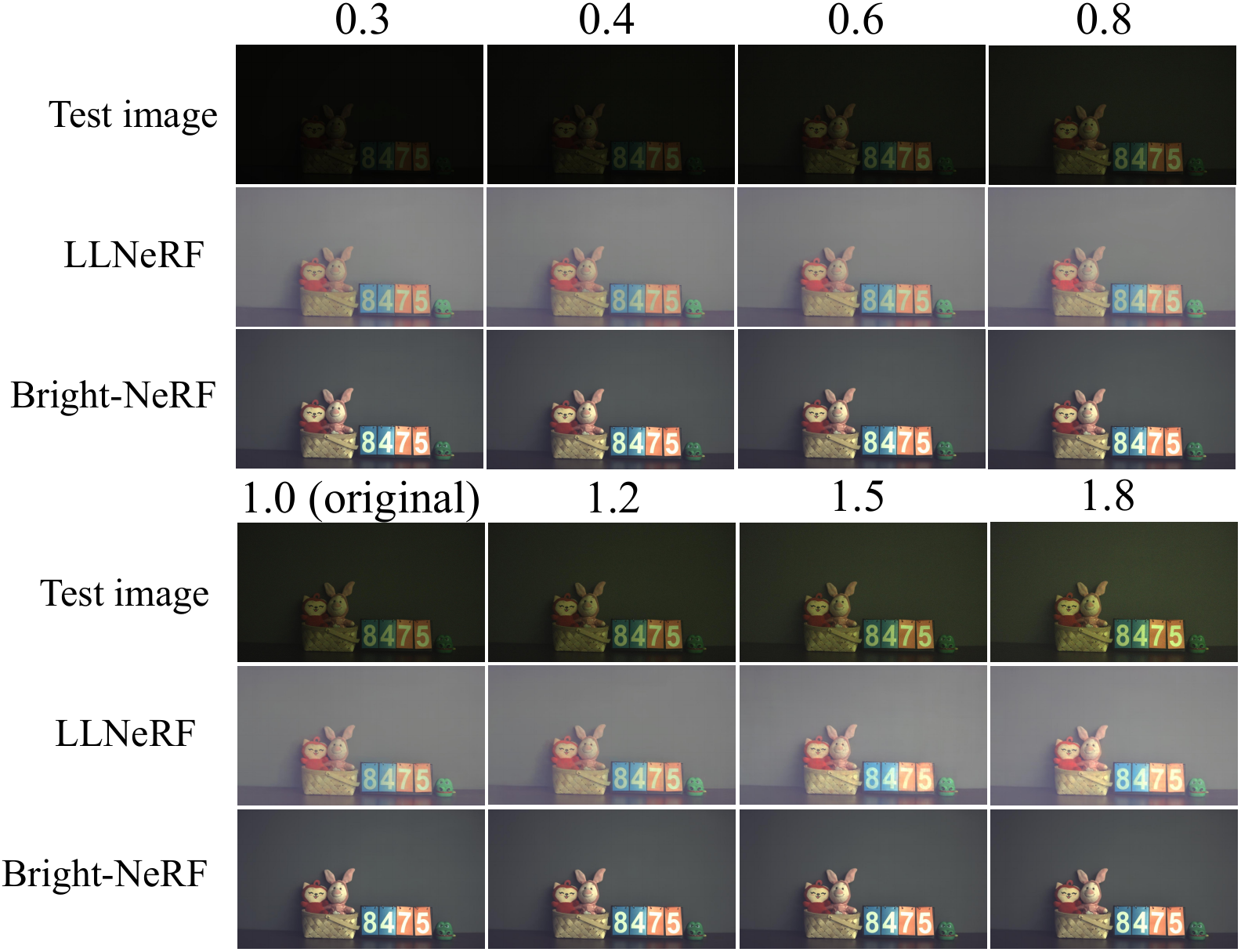}
\caption{
\textbf{Performance comparison on scenes with different darkness degrees.}
As the simulated exposure ratio decreases (darker scene), LLNeRF's results exhibit more severe color bias while our Bright-NeRF demonstrates adaptability and stability to different degrees of darkness, producing natural and stable color even in extremely low-light conditions.}
\label{appendFig.1}
\end{figure*}

\begin{figure*}
\centering
\includegraphics[width=1\textwidth]{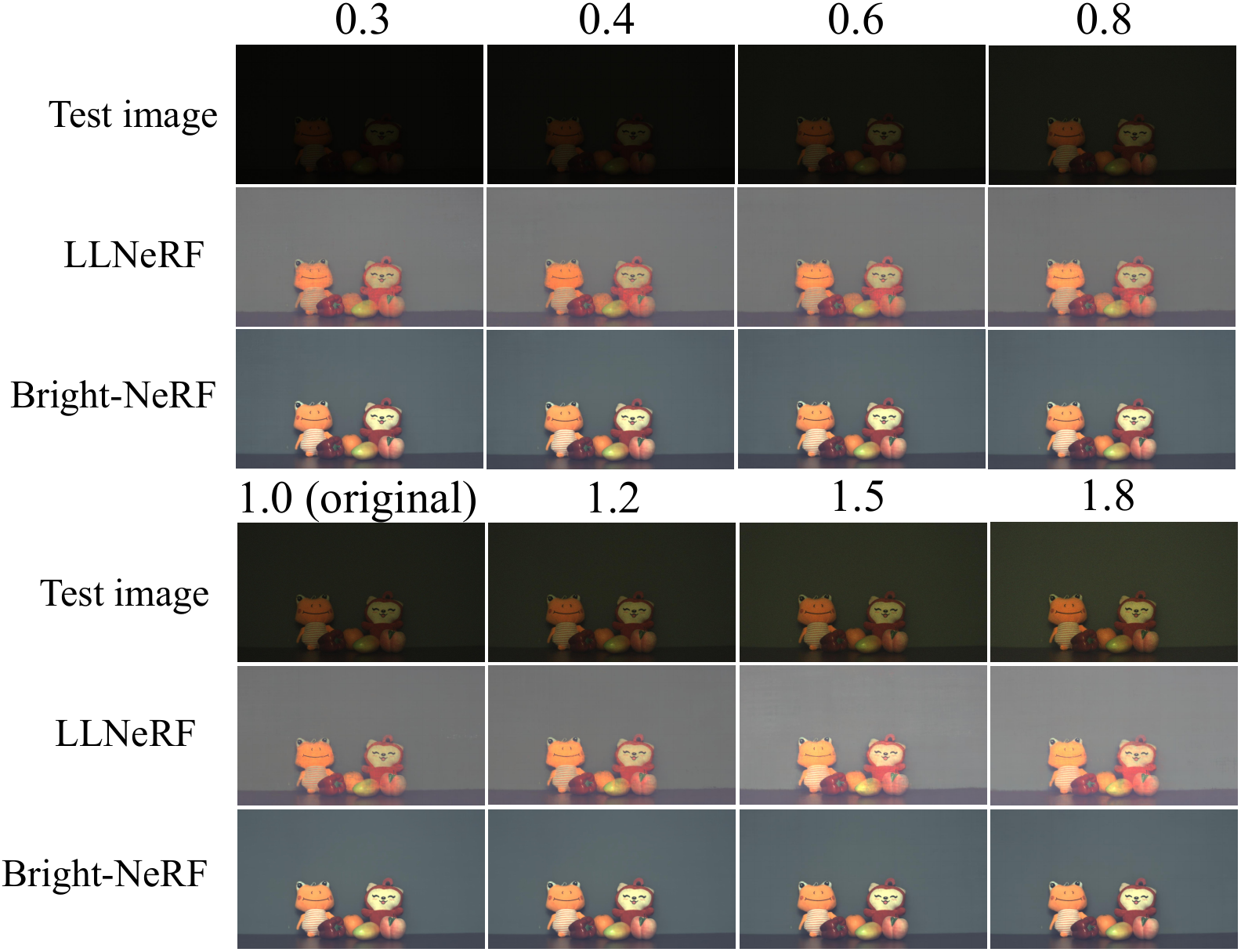}
\caption{
\textbf{Performance comparison on scenes with different darkness degrees.}
As the simulated exposure ratio decreases (darker scene), LLNeRF's results exhibit more severe color bias while our Bright-NeRF demonstrates adaptability and stability to different degrees of darkness, producing natural and stable color even in extremely low-light conditions.}
\label{appendFig.2}
\end{figure*}

We set a series of exposure ratios to simulate different darkness degrees to evaluate the performance of our method under different lighting conditions. Inspired by SID \cite{chen2018learning} and EEMEFN \cite{zhu2020eemefn}, the former defines the exposure ratio $\gamma^*$ as the exposure differences between the low-light image and the reference image, while the latter uses $\left\lbrace 1, \gamma^* \right\rbrace$ as the exposure ratios to simulate multi-exposure images. These methods leverage the linear relationship between signals and the number of received photons in the raw domain, amplifying the intensity of raw signals to expose the low-light images. We also leverage this characteristic but rather to simulate both darker and brighter than current raw images by setting the exposure ratio to $\gamma$ times the current exposure time ($ \gamma \in \left\lbrace 0.3, 0.4, 0.6, 0.8, 1.0, 1.2, 1.5, 1.8 \right \rbrace$). 
\begin{equation}
    I_{\gamma} = I_{raw} \cdot \gamma, 
\end{equation}
where $I_{raw} \in \mathbb{R}^{H \times W \times 1}$ is the mosaiced raw image after subtracting the black level and $I_{\gamma}$ is the scaled linear data by exposure ratio $\gamma$. More comparisons between LLNeRF \cite{wang2023lighting} and our Bright-NeRF under different lighting conditions are illustrated in Fig. \ref{appendFig.1} and Fig. \ref{appendFig.2}.

\section{Additional Results}
\begin{figure*}[h]
\centering
\includegraphics[width=1\textwidth]{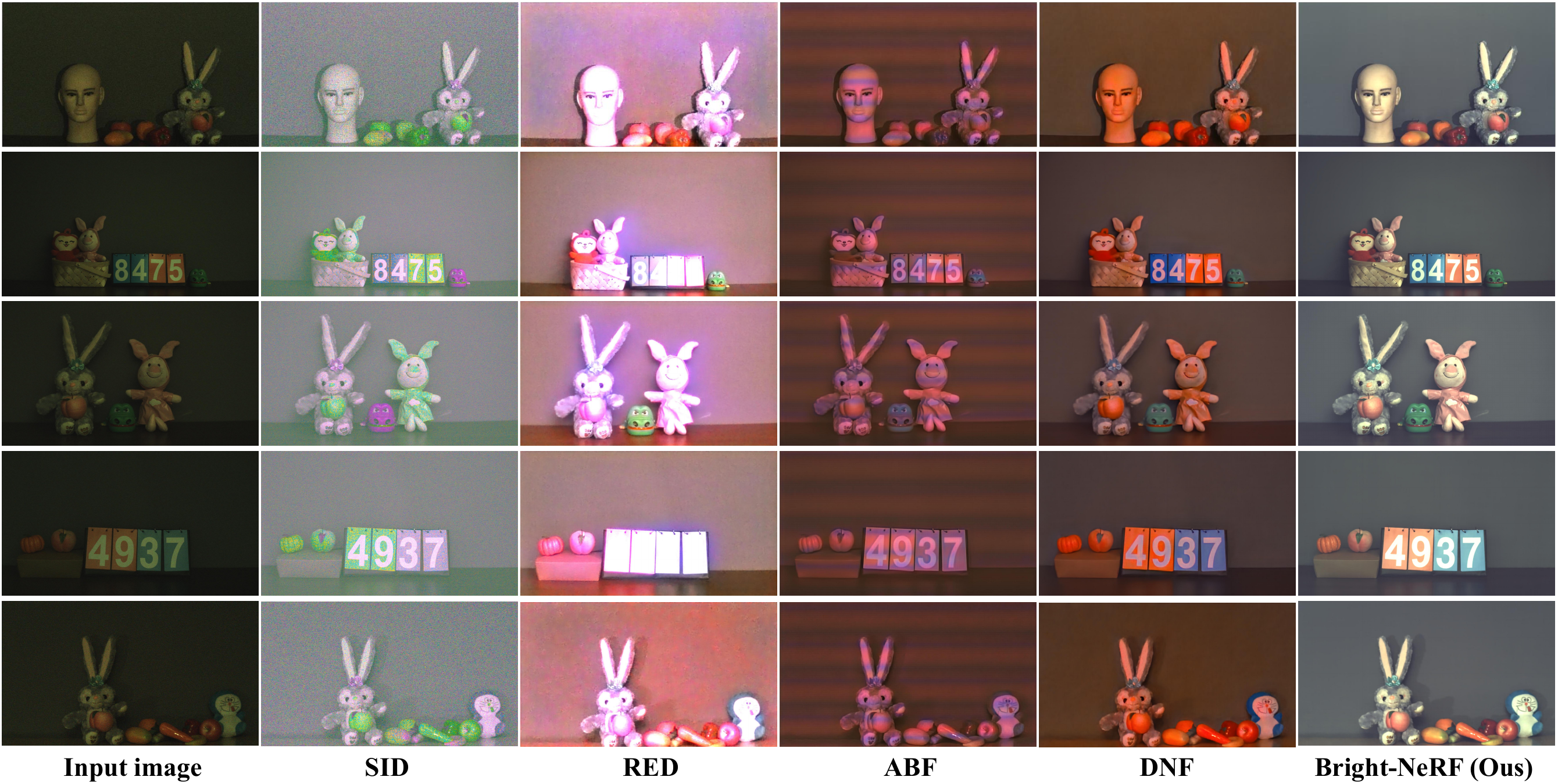} 
\caption{{\textbf{Qualitative comparison with state-of-the-art 2D LLIE methods.}} We show that our method effectively recovers detailed geometry and accurate colors while other methods often struggle with denoising and tend to produce incorrect colors. For these 2D methods, we use publicly available code and pre-trained model weights. }
\label{appendFig.3}
\end{figure*}

\begin{figure*}[h]
\centering
\includegraphics[width=1.0\textwidth]{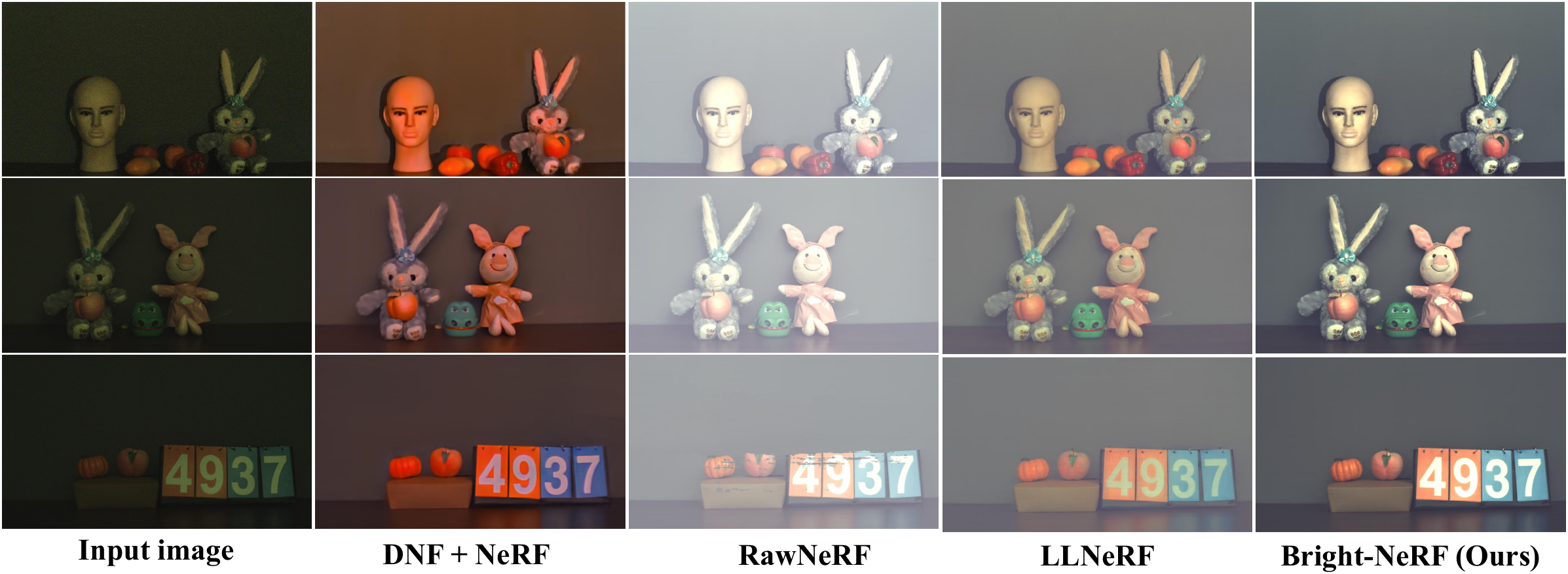} 
\caption{{\textbf{Qualitative comparison with state-of-the-art LLIE NeRF methods.}} Leveraging the sensor's adaptive responses in the color restoration field, the results of our Bright-NeRF exhibit more natural color and more accurate tone restoration.}
\label{appendFig.4}
\end{figure*}

We provide additional quantitative and qualitative comparisons with 2D LLIE methods and LLIE NeRF methods. Tab. \ref{appendTab.1} includes a detailed breakdown of the quantitative results presented in the main paper into per-scene metrics. To compare with the sRGB-based NeRF method (\textit{e.g.} LLNeRF \cite{wang2023lighting}), we compute the evaluation metrics in sRGB domain. The quantitative results further validate that our approach outperforms existing state-of-the-art methods. Fig. \ref{appendFig.3} and Fig. \ref{appendFig.4} illustrate the qualitative results of our method and existing methods, showing that our approach enhances brightness effectively while restoring more vivid and natural colors. 

As shown in Fig. \ref{appendFig.3}, compared to existing 2D methods, which tend to overfit to the training data processed by a fixed ISP, leading to poor generalization across different cameras and illumination conditions, our Bright-NeRF can adaptively correct color distortion and denoise without supervision.

As illustrated in Fig. \ref{appendFig.4}, the baseline model’s results show noticeable discrepancies in color reproduction which exhibits incorrect color. While RawNeRF \cite{mildenhall2022nerf} successfully increases the brightness, it fails to produce a natural appearance. The renderings of LLNeRF \cite{wang2023lighting} show insufficient restoration of tone and saturation. Compared to the above methods, our method takes full use of the characteristics of raw data, enabling the recovery of inherent scene colors even in extremely low-light conditions. Additionally, our method does not rely on a pre-calibrated ISP for pre- or post-processing views, thereby reducing the complexity of ISP calibration while avoiding the sub-optimal results often associated with generic ISPs. Our method achieves \textbf{end-to-end ISP-free} synthesis of novel views under normal lighting conditions from a set of low-light data.

We further compare the performance of our method with LLNeRF \cite{wang2023lighting} by training both models on sRGB images. The results in Table. \ref{appendTab.2} highlight the domain-stability advantage of our Bright-NeRF across raw and sRGB domain.

\begin{table}[h]

\centering
\begin{tabular}{ l|cccc } 

 \hline
 \hline
   Method  &  PSNR $\uparrow$ & SSIM $\uparrow$ & LPIPS $\downarrow$ & LIQE  $\uparrow $\\
 \hline

  LLNeRF  & 16.64 & 0.811 & 0.494 & 1.332
  \\
  \hline
  Ours &  \textbf{21.49} & \textbf{0.868} & \textbf{0.477} & \textbf{1.562} 
  \\

\hline
\hline 
\end{tabular}

\caption{\textbf{Quantitative comparison with LLNeRF on sRGB images.} The calculated metrics are the average values across nine scenes. The best results are marked in {\bf bold}.} 
 \label{appendTab.2}
\end{table}

\begin{figure*}[h]
\centering
\includegraphics[width=1.0\textwidth]{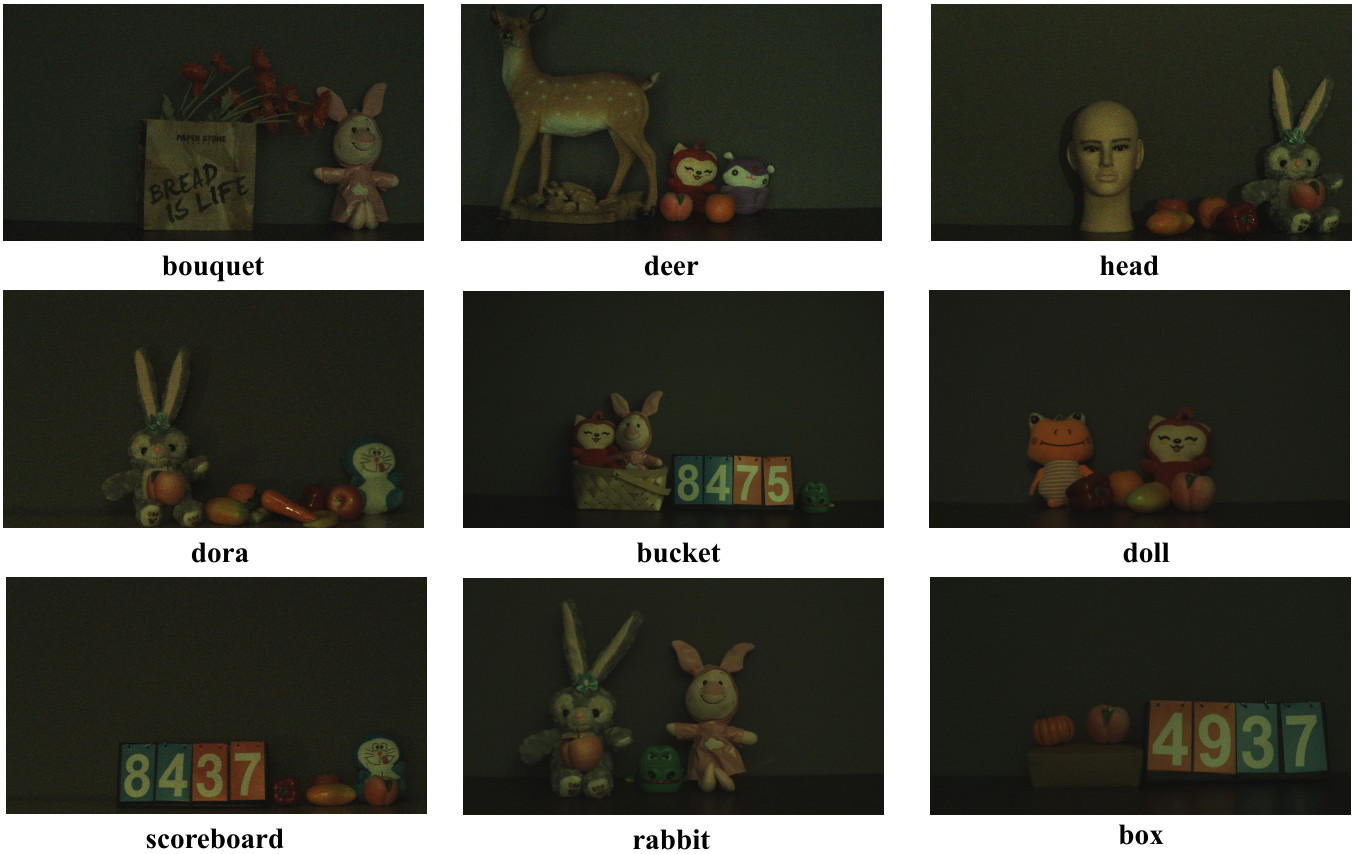} 
\caption{Illustration of example images of scenes from our proposed dataset.}
\label{appendFig.5}
\end{figure*}

\begin{table*}[h]
\centering
\begin{tabular}{l|c|cccc|cccc}
\hline
\hline
\multirow{2}{*}{Method} & \multirow{2}{*}{Method Type} & \multicolumn{4}{c|}{\textit{bouquet}} & \multicolumn{4}{c}{\textit{deer}} \\
\cline{3-10}
 && PSNR$\uparrow$ & SSIM$\uparrow$ & LPIPS $\downarrow$ & LIQE $\uparrow$ & PSNR$\uparrow$ & SSIM$\uparrow$ & LPIPS $\downarrow$ & LIQE $\uparrow$  \\
\hline
SID \cite{chen2018learning}
& \multirow{4}{*}{2D} & 13.40 & 0.135 & 0.859 & 1.001 & 15.91 & 0.141 & 0.865 & 1.001 \\

RED \cite{lamba2021restoring} && 12.93 & 0.674 & 0.686 & 1.001 & 12.24 & 0.702 & 0.688 & 1.000 \\

ABF \cite{dong2022abandoning} & &16.78 & 0.798 & 0.672 & 1.003 & 15.30 & 0.807 & 0.681 & 1.002\\

DNF \cite{jin2023dnf} && 17.02 & 0.814 & 0.603 & 1.108 & 15.70 & 0.824 & 0.600 & 1.071  \\
\hline
RawNeRF \cite{mildenhall2022nerf} & \multirow{3}{*}{3D} & 7.96 & 0.669 & 0.556 & 1.271 & 10.46 & 0.777 & 0.508  &1.323\\

DNF \cite{jin2023dnf} + NeRF && 16.61 & 0.801 & 0.635 & 1.096 &15.566 & 0.819 &0.621 & 1.079  \\

LLNeRF \cite{wang2023lighting} && 17.55 & 0.816 & 0.502 & 1.163 & 21.42 & 0.871 & 0.498 &1.128\\
\hline
Ours & 3D & \textbf{19.07} & \textbf{0.844} & \textbf{0.483} & \textbf{1.391} & \textbf{22.59} & \textbf{0.889} & \textbf{0.449} & \textbf{1.487} \\
\hline \hline

\multirow{2}{*}{Method} & \multirow{2}{*}{Method Type} & \multicolumn{4}{c|}{\textit{head}} & \multicolumn{4}{c}{\textit{rabbit}} \\
\cline{3-10}
 && PSNR$\uparrow$ & SSIM$\uparrow$ & LPIPS $\downarrow$ & LIQE $\uparrow$ & PSNR$\uparrow$ & SSIM$\uparrow$ & LPIPS $\downarrow$ & LIQE $\uparrow$  \\
\hline
SID \cite{chen2018learning}
& \multirow{4}{*}{2D} & 15.67 & 0.149 & 0.845 & 1.001 & 12.91 & 0.381 & 0.692 & 1.007 \\

RED \cite{lamba2021restoring} && 10.95 & 0.652 & 0.761 & 1.000 & 9.62 & 0.601 & 0.761 & 1.001 \\

ABF \cite{dong2022abandoning} & &15.13 & 0.794 & 0.705 & 1.003 & 18.98 & 0.836 & 0.683 & 1.003\\

DNF \cite{jin2023dnf} && 16.00 & 0.823 & 0.710 & 1.090 & 19.99 & 0.854 & 0.612 & 1.091  \\
\hline
RawNeRF \cite{mildenhall2022nerf} & \multirow{3}{*}{3D} & 10.40 & 0.771 & 0.517 & 1.249 & 8.28 & 0.664 & 0.518  &1.311\\

DNF \cite{jin2023dnf} + NeRF && 15.93 & 0.819 & 0.733 & 1.072 &19.92 & 0.844 &0.632 & 1.072  \\

LLNeRF \cite{wang2023lighting} && 20.77 & 0.865 & 0.494 & 1.192 & 15.51 & 0.799 & 0.496 &1.241\\
\hline
Ours & 3D & \textbf{24.05} & \textbf{0.885} & \textbf{0.474} & \textbf{1.356} & \textbf{20.38} & \textbf{0.862} & \textbf{0.482} & \textbf{1.404} \\
\hline \hline

\multirow{2}{*}{Method} & \multirow{2}{*}{Method Type} & \multicolumn{4}{c|}{\textit{bucket}} & \multicolumn{4}{c}{\textit{doll}} \\
\cline{3-10}
 && PSNR$\uparrow$ & SSIM$\uparrow$ & LPIPS $\downarrow$ & LIQE $\uparrow$ & PSNR$\uparrow$ & SSIM$\uparrow$ & LPIPS $\downarrow$ & LIQE $\uparrow$  \\
\hline
SID \cite{chen2018learning}
& \multirow{4}{*}{2D} & 12.48 & 0.387 & 0.694 & 1.011 & 12.03 & 0.343 & 0.703 & 1.010 \\

RED \cite{lamba2021restoring} && 10.69 & 0.660 & 0.678 & 1.007 & 9.64 & 0.605 & 0.710 & 1.005 \\

ABF \cite{dong2022abandoning} & &19.70 & 0.850 & 0.644 & 1.015 & 19.80 & 0.840 & 0.645 & 1.008\\

DNF \cite{jin2023dnf} && 20.75 & 0.866 & 0.552 & 1.311 & 20.73 & 0.854 & 0.557 & 1.134  \\
\hline
RawNeRF \cite{mildenhall2022nerf} & \multirow{3}{*}{3D} & 8.81 & 0.680 & 0.492 & 2.171 & 8.17 & 0.642 & 0.496  &1.785\\

DNF \cite{jin2023dnf} + NeRF && 20.51 & 0.858 & 0.565 & 1.287 &20.44 & 0.845 &0.569 & 1.114  \\

LLNeRF \cite{wang2023lighting} && 14.61 & 0.786 & 0.468 & 1.541 & 13.97 & 0.750 & 0.479 &1.301\\
\hline
Ours & 3D & \textbf{25.26} & \textbf{0.887} & \textbf{0.456} & \textbf{2.227} & \textbf{22.12} & \textbf{0.861} & \textbf{0.451} & \textbf{1.786} \\
\hline \hline
\multirow{2}{*}{Method} & \multirow{2}{*}{Method Type} & \multicolumn{4}{c|}{\textit{dora}} & \multicolumn{4}{c}{\textit{scoreboard}} \\
\cline{3-10}
 && PSNR$\uparrow$ & SSIM$\uparrow$ & LPIPS $\downarrow$ & LIQE $\uparrow$ & PSNR$\uparrow$ & SSIM$\uparrow$ & LPIPS $\downarrow$ & LIQE $\uparrow$  \\
\hline
SID \cite{chen2018learning}
& \multirow{4}{*}{2D} & 13.37 & 0.140 & 0.849 & 1.001 & 12.62 & 0.127 & 0.852 & 1.001 \\

RED \cite{lamba2021restoring} && 12.07 & 0.672 & 0.784 & 1.001 & 11.02 & 0.657 & 0.807 & 1.004 \\

ABF \cite{dong2022abandoning} & &17.57 & 0.819 & 0.710 & 1.002 & 18.41 & 0.829 & 0.699 & 1.029\\

DNF \cite{jin2023dnf} &&18.16 & 0.831 & 0.771 & 1.111 &18.617 & 0.831 & 0.771 & 1.525  \\
\hline
RawNeRF \cite{mildenhall2022nerf} & \multirow{3}{*}{3D} & 8.97 & 0.704 & 0.544 & 1.254 & 9.38 & 0.702 & 0.537  &2.339\\

DNF \cite{jin2023dnf} + NeRF && 17.95 & 0.823 & 0.800 & 1.104 &16.86 & 0.789 &0.841 & 1.238  \\

LLNeRF \cite{wang2023lighting} && 16.85 & 0.831 & 0.511 & 1.092 & 15.07 & 0.808 & 0.507 &1.785\\
\hline
Ours & 3D & \textbf{24.76} & \textbf{0.885} & \textbf{0.505} & \textbf{1.128} & \textbf{24.72} & \textbf{0.887} & \textbf{0.503} & \textbf{2.144} \\

\hline \hline
\multirow{2}{*}{Method} & \multirow{2}{*}{Method Type} & \multicolumn{4}{c|}{\textit{box}} & \multicolumn{4}{c}{\textit{mean}} \\
\cline{3-10}
 && PSNR$\uparrow$ & SSIM$\uparrow$ & LPIPS $\downarrow$ & LIQE $\uparrow$ & PSNR$\uparrow$ & SSIM$\uparrow$ & LPIPS $\downarrow$ & LIQE $\uparrow$  \\
\hline
SID \cite{chen2018learning}
& \multirow{4}{*}{2D} & 11.95 & 0.299 & 0.714 & 1.001 & 13.37 & 0.233 & 0.786 & 1.004 \\

RED \cite{lamba2021restoring} && 10.18 & 0.658 & 0.701 & 1.006 & 11.04 & 0.654 & 0.731 & 1.003 \\

ABF \cite{dong2022abandoning} & &19.98 & 0.861 & 0.637 & 1.025 & 17.96 & 0.826 & 0.675 & 1.010\\

DNF \cite{jin2023dnf} &&20.40 & 0.871 & 0.547 & 1.456 &18.60 & 0.841 & 0.636 & 1.211  \\
\hline
RawNeRF \cite{mildenhall2022nerf} & \multirow{3}{*}{3D} & 8.07 & 0.636 & 0.546 & 1.932 & 8.95 & 0.694 & 0.524  &1.626\\

DNF \cite{jin2023dnf} + NeRF && 20.04 & 0.862 & 0.568 & 1.299 &18.20 & 0.829 &0.663 & 1.151  \\

LLNeRF \cite{wang2023lighting} && 13.99 & 0.777 & 0.493 & 1.550 & 16.64 & 0.811 & 0.494 &1.332\\
\hline
Ours & 3D & \textbf{20.82} & \textbf{0.870} & \textbf{0.478} & \textbf{2.009} & \textbf{22.64} & \textbf{0.874} & \textbf{0.476} & \textbf{1.659} \\
\hline \hline

\end{tabular}
\caption{\textbf{Quantitative comparison with existing LLIE methods.} We provide a per-scene breakdown of the quantitative results presented in the main paper. The best results are marked in {\bf bold}.}
\label{appendTab.1}
\end{table*}

\end{document}